\documentclass[sigconf]{aamas} 
\setcopyright{none}
\acmConference[noacm]{}
\acmDOI{}
\acmPrice{}
\acmISBN{}
\acmPrice{}
\usepackage{balance} % for balancing columns on the final page
\usepackage[frozencache,cachedir=.]{minted}
\title[GAMMS]{GAMMS: Graph based Adversarial Multiagent Modeling Simulator}

\author{Rohan Patil}
\authornote{Both authors contributed equally to the paper}
\affiliation{
  \institution{UC San Diego}
  \city{San Diego}
  \country{United States}}
\email{rpatil@ucsd.edu}

\author{Jai Malegaonkar}
\authornotemark[1]
\affiliation{
  \institution{UC San Diego}
  \city{San Diego}
  \country{United States}}
\email{jmalegaonkar@ucsd.edu}

\author{Xiao Jiang}
\authornote{Both authors contributed equally to the paper}
\affiliation{
  \institution{University of Southern California}
  \city{Los Angeles}
  \country{United States}}
\email{xjiang26@usc.edu}

\author{Andre Dion}
\authornotemark[2]
\affiliation{
  \institution{University of Southern California}
  \city{Los Angeles}
  \country{United States}}
\email{adion@usc.edu}

\author{Gaurav S. Sukhatme}
\affiliation{
  \institution{University of Southern California}
  \city{Los Angeles}
  \country{United States}}
\email{gaurav@usc.edu}

\author{Henrik I. Christensen}
\affiliation{
  \institution{UC San Diego}
  \city{San Diego}
  \country{United States}}
\email{hichristensen@ucsd.edu}

\begin{abstract}
As intelligent systems and multi-agent coordination become increasingly central to real-world applications, there is a growing need for simulation tools that are both scalable and accessible. Existing high-fidelity simulators, while powerful, are often computationally expensive and ill-suited for rapid prototyping or large-scale agent deployments. We present GAMMS (Graph based Adversarial Multiagent Modeling Simulator), a lightweight yet extensible simulation framework designed to support fast development and evaluation of agent behavior in environments that can be represented as graphs. GAMMS emphasizes five core objectives: scalability, ease of use, integration-first architecture, fast visualization feedback, and real-world grounding. It enables efficient simulation of complex domains such as urban road networks and communication systems, supports integration with external tools (e.g., machine learning libraries, planning solvers), and provides built-in visualization with minimal configuration. GAMMS is agnostic to policy type, supporting heuristic, optimization-based, and learning-based agents, including those using large language models. By lowering the barrier to entry for researchers and enabling high-performance simulations on standard hardware, GAMMS facilitates experimentation and innovation in multi-agent systems, autonomous planning, and adversarial modeling. The framework is open-source and available at \url{https://github.com/GAMMSim/GAMMS/}
\end{abstract}

\keywords{Robot Planning, Simulation, Multi-agent}

\newcommand{\BibTeX}{\rm B\kern-.05em{\sc i\kern-.025em b}\kern-.08em\TeX}

\begin{document}

%%% The following commands remove the headers in your paper. For final 
%%% papers, these will be inserted during the pagination process.

\pagestyle{fancy}
\fancyhead{}

%%% The next command prints the information defined in the preamble.

\maketitle 

%%%%%%%%%%%%%%%%%%%%%%%%%%%%%%%%%%%%%%%%%%%%%%%%%%%%%%%%%%%%%%%%%%%%%%%%

\section{Introduction}
\label{sec:introduction}

    % Challenge of developing complex planning algorithms in robotics
    Robot automation has moved beyond simple automation in structured industrial settings~\cite{bahrin2016industry} to serving coffee to humans~\cite{sung2020untact}, guide around an exhibit~\cite{iio2020human} as well as helping to clean homes~\cite{forlizzi2006service}. A major part of these intelligent robot systems to navigate and understand the complexities of our world is thanks to the sophisticated planning algorithms that convert high level instructions into low level movement logic for the robot to execute. However, it becomes increasingly difficult to test the behavior of the planning algorithm in various situations in the real world due to physical constraints. In the prototype phase, usually the goals are to test a particular part of the system. However, the planning system is dependent on the proper functioning of the control logic which further depends on proper functioning of the various sensors and mechanical components. Simulation provides an efficient way to test the basic behavior of the planning system. Obviously, simulation is not a perfect solution but it provides quick initial feedback to  iterate during prototyping phase. This is especially helpful in the current approaches that utilize deep neural networks. Due to the black box nature of these approaches, there are multiple scenarios where it is necessary to test the behavior of the system and validate it. In addition, heuristics are an important engineering tool but need to be tested to get an idea of its efficacy.
    
    % Complexity and scaling issue of end-to-end high fidelity simulators
    In an ideal world, the simulation exactly simulates every part of the robot system as well as the environment. However, the compute cost increases based on the detail to which everything is simulated. There is a direct relation between the fidelity of the simulation and its scalability with compute. The scalability issue is further extenuated when dealing with system that have multiple robots. The scalability problem highlights the limit of using high fidelity for testing multi-robot systems in prototyping phase as well as in scenarios where the motive is to observe a multi-robot system's overall high level plan or strategy.
    
    % Separation of control and planning (Write this with context of new developments of highlevel planning using LLMs)
    In contrast to end-to-end systems, the recent development of agentic systems have shown that it is possible to develop components independently and finally put them together to create a highly intelligent systems. The upcoming field of embodied artificial intelligence is relying on large language models to do a high level planning for tasks while using a mix of classical control or reinforcement learning based control to create the low level instructions. In addition, it has been observed that LLMs have many limitations when generating low-level (control) instructions accurately~\cite{kim2024survey}. This separation of control and high-level planning creates the opportunity to develop a simulator directed towards testing the behavior using abstract data that have some grounding with the real world but need not be exact~\cite{truong2023rethinking, noorani2025abstraction, labiosa2025reinforcement}.
    
    %Gamms addresses a specific niche of developing strategies on systems (like road-networks) that can be abstracted as a graph( the other niche is 1000+ nodes )
    In this paper, we present GAMMS (Graph based Adversarial Multiagent Modeling Simulator), a simulator targeted towards addressing the issue of testing and developing strategies for systems that can be abstracted as a graph structure. Graph-based abstraction is ideal for large real-world environments because it efficiently models complex relationships and dependencies between entities, enabling scalable analysis and decision-making. It allows for flexible navigation, optimization, and problem-solving across dynamic and interconnected systems. Road networks are a prime example of real-world systems that can be represented as a graph. In the case of autonomous cars or robots in general, the road connectivity encodes the possible routes the agent can take to travel from one point to another. However, it should be noted that such graphs can be large (1000+ nodes). In the city of Manhattan itself, there are over 2800 traffic signals~\cite{nycInfrastructureTraffic}. Another reason for developing GAMMS is to have a structure that helps make it possible to simulate at scale for such large graphs.

    \begin{figure}[h]
        \centering
        \includegraphics[width=\linewidth]{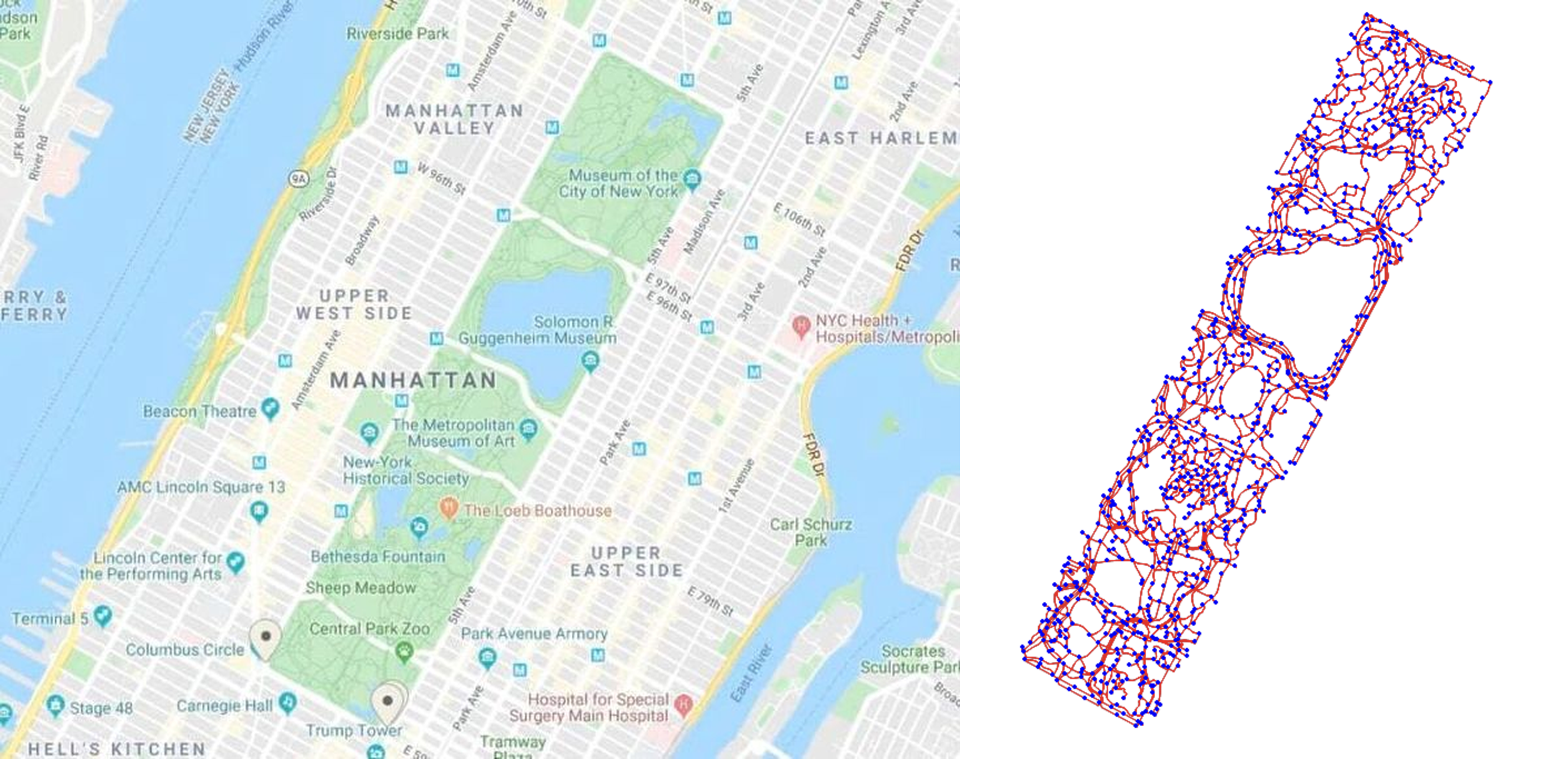}
        \caption{Central Park, NYC OSM on the left. GAMMS graph for Central Park on the right.}
        \label{fig:central_park}
    \end{figure}
    % (Include figures of OSM maps loaded into gamms as teaser)
    % \todo[inline]{Extend more on why graphs are a good abstraction. Add figures. Figure of Manhattan OSM roads and the converted GAMMS graph snapshot for comparison.}
\section{GAMMS Objectives}
\label{sec:objectives}
The design and development of GAMMS is influenced by five core objectives that address fundamental challenges/issues in multi-agent simulation research: 1. Scalability, 2. Ease of Use,  3. Integration First Arch, 4. Fast Visualization Feedback, 5. Real World Grounding.  % rohan u can shop this later
Through these objectives we intend to democratize access to large-scale  multi-agent modeling while maintaining the flexibility needed for diverse research applications.

\textbf{Scalable Graph-Based Simulation.} GAMMS prioritizes computational efficiency to enable large scale multi-agent scenarios that can run on standard consumer hardware rather than requiring specialized computing resources. The graph based environmental representation provides a natural abstraction for many real-world domains where agents operate on network structures such as urban road systems, communication networks, and social interaction graphs. The simulator implements a simple interface which does not enforce any particular semantics, providing flexibility for researchers working with different paradigms. The ability to scale simulation to large graphs (1000+ nodes) allows researchers to simulate multi-agent systems on a scale similar to the real world while maintaining reasonable computational performance. This addresses a key barrier that often limits multi-agent research to small-scale toy problems.

\textbf{Ease of Use.} A critical objective of GAMMS is to reduce the barrier to entry for researchers who want to experiment with multi-agent scenarios but lack extensive software engineering experience. The simulator is designed to require minimal training and setup time, allowing domain experts to quickly translate their conceptual models into running simulations. We recognize that many valuable insights in multi-agent research come from researchers whose primary expertise lies in application domains (economics, planning, game theory, etc.) rather than computer science, and that complex simulation frameworks often prevent these researchers from effectively testing their hypotheses. 
% IDK add some thing about Daigo work or Shaunak.

\textbf{Integration-First Architecture.} The framework is explicitly designed with an \textit{integration-first} philosophy, making it easier to incorporate external tools, libraries, and existing code rather than requiring users to re-implement functionality within a closed ecosystem. This architectural decision acknowledges that effective multi-agent research often requires combining insights and tools from multiple domains. For example, deep/machine/reinforcement learning libraries for agent intelligence, optimization solvers for planning problems, statistical packages for analysis, or domain-specific simulators for particular environmental effects. By prioritizing integration capabilities, GAMMS enables researchers to leverage existing tools while focusing their efforts on the novel aspects of their research.
% Maybe combine with prev section

\textbf{Fast Visualization Feedback.} Building on top of the previous two objectives, GAMMS also emphasizes minimal code requirements for creating meaningful visualizations of simulation results. Rather than requiring users to implement complex rendering pipelines or learn specialized graphics libraries, the framework provides built-in visualization capabilities that can display agent behaviors, environmental states, and simulation dynamics with none to minimal additional code. This objective stems from the recognition that visualization is crucial for both debugging multi-agent systems and communicating results.

\textbf{Real-World Grounding.} GAMMS is designed to facilitate the modeling of scenarios that connect to real-world data and environments, rather than focusing exclusively on abstract or toy problems. The framework includes native support for processing real-world geographical data (such as OpenStreetMap~\cite{OpenStreetMap}) and converting it into graph representations suitable for multi-agent simulation. This objective reflects the understanding that many of the most important questions in multi-agent systems research ultimately concern how agents behave in realistic environments with real constraints, and that simulators should facilitate this connection to reality rather than abstracting away from it.

These objectives collectively position GAMMS as a practical tool for advancing multi-agent systems research by lowering barriers to experimentation while maintaining the flexibility and realism needed for meaningful scientific investigation.
\section{Related Works}
\label{sec:related_works}
    Computer simulation has grown hand-in-hand with the rapid growth in computing power. In the context of multi-agent simulators, multiple simulators to test simple automatons like in Conway's Game of Life have been made. As such, there is a large body of work here to consider, and hence we limit our discussion primarily to simulators that have been used in the context of developing multi-robot systems. We divide the simulators into two groups: high-fidelity and low-fidelity simulators. High-fidelity simulators aim to closely replicate real-world conditions, incorporating detailed features such as realistic physics models, non-idealized sensor data, and real-time feedback mechanisms. A key advantage of high-fidelity environments is that algorithms validated within these simulators tend to perform reliably on physical robots. In contrast, low-fidelity simulators offer simplified abstractions of reality, emphasizing lightweight codebases that facilitate rapid prototyping and enable extensive parallel experimentation. These environments are geared towards high-level algorithm development, allowing researchers to quickly iterate on new ideas before moving to more complex testing phases. This distinction is particularly relevant in the context of GAMMS and related frameworks, where balancing fidelity with development efficiency is crucial.

    \textbf{High Fidelity Simulators.} High-fidelity simulation frameworks serve as sophisticated tools for constructing realistic simulations. They provide modular architectures that allow users to compose physics, sensors, environments, and execution pipelines flexibly. For example, Gazebo and Isaac Sim~\cite{liang2018gpu} integrate tightly with ROS2~\cite{doi:10.1126/scirobotics.abm6074} through a ROS Bridge, enabling simulation components to publish and subscribe to standard ROS topics such as joint states, camera images, and transforms in real time. They support a broad range of realistic sensors sourced from industry manufacturers (e.g., Intel, Orbbec, SICK) and allow users to implement custom sensors~\cite{rishabh2024beginners}. Isaac Sim leverages GPU-accelerated ray-traced rendering and physics simulation to deliver highly realistic environments and sensor data. Such frameworks support importing robot models via URDF or USD formats, scene composition with complex objects, and precise physical interactions. However, using these frameworks as tools requires significant setup effort, advanced hardware (e.g., RTX GPUs for Isaac Sim), and expertise to tune simulation parameters. Iteration can be slow due to high computational demands and complexity, especially when scaling to large numbers of agents or structured multi-team scenarios. Consequently, while high-fidelity tools excel at final validation, robot-in-the-loop testing, and sim-to-real transfer, their overhead can be prohibitive during early-stage algorithm development.

    \textbf{Low Fidelity Simulators.}  Low-fidelity simulators prioritize simplicity to enable rapid iteration and better scalability during early development stages. Designed primarily for prototyping and early algorithm testing, they rely on idealized sensors, simplified physics, and minimal computational overhead, allowing developers to focus on high-level logic and strategy rather than intricate low-level details. However, this simplicity often results in behaviors that diverge significantly from real-world dynamics, limiting their direct applicability to production environments. Common agent-based modeling frameworks like Repast~\cite{North2013}, Mason~\cite{Luke2005MASON}, Agents.jl~\cite{Agents.jl}, and GAMA~\cite{Amouroux2009} offer greater flexibility in spatial, network, and agent interactions and can handle larger populations, but they are not Python-native and typically it becomes harder to integrate with other existing software. NetLogo~\cite{Wilensky:1999}, which can be integrated with Python for parallel and headless execution, offers ease of use but is not optimized for handling massive agent populations or large environments. Mesa~\cite{terHoeven2025} is fully Python-based and supports both grid and continuous environments, with extensions for hierarchical agent groupings. It was created in response to the complexity of previous simulators to leverage the Python ecosystem similar to GAMMS. However, the primary focus of Mesa is on modeling artificial societies. In general, we do not find a big intersection between robotics and low fidelity simulators.

\section{Simulator Architecture}
\label{sec:simulator_architecture}

    In this section, we will cover the design principles that GAMMS adheres to in its development. We will also briefly discuss why the particular choice was made. Followed by that, we will describe the internal design of the simulator based on the design principles and cover briefly how it helps achieve the objectives mentioned in the previous section. Finally, we conclude the section with an overview of the overall API that a user gets to develop scenarios using GAMMS.
    
    \subsection{Design Principles and Choices}   
    \label{subsec:design_principles}
        We will first discuss the design principles followed by the particular choices made during the actual implementation of the simulator. 
    
        \textbf{Minimal Maintained Interfaces.} ``Everything is a file'' is a well-known concept in Linux philosophy~\cite{yarchiveEverythingisafilePrinciple}. Having a fewer number of interfaces makes it easier for the tool user to implement as well as extend functionality. As there are only a few things the user needs to know to get started, it helps ensure achieving the ``Ease of Use'' objective. From a testing perspective, it becomes easier to write unit tests for correctness~\cite{beyer2020interface}. Furthermore, for integrating other tools, a developer at most might need to re-implement the interfaces. Hence, keeping a small number of interfaces helps in integration with external tools.

        % Cite for interface verifiability https://link.springer.com/content/pdf/10.1007/978-3-030-61362-4_9.pdf

        \textbf{Stable Interface Contract.} Only limiting the number of interfaces does not ensure ease in developing and extending the implementations for integrating other tools or pieces of code in general. While many rely on modularity and try to maintain loose coupling of different components, GAMMS accepts the fact that such implementations are time-consuming to develop and put many restrictions on development. Instead, GAMMS takes the ``Design by Contract''~\cite{meyer2002applying} philosophy for each maintained interface. Since the contracts are formalized, it is easier to reason about the behavior of components when maintaining or extending the software. Furthermore, it becomes possible to maintain a minimum level of documentation that is \textit{complete} for the user to use the tool.

        \textbf{Intermediate Representation when Unstructured.} There are certain components that are internal to the development work and do not have a maintained interface. An example in GAMMS is the recording module that emits an optimized representation of the entire sequence of states that happens during a scenario run. What and how the recording happens depends on the implementation in question, hence it becomes impossible to maintain a strict contract without compromising on integration ease. The recorder instead uses maintained representations which provide flexibility to work in an implementation-independent format. This is inline with the concept of transparent runtime model~\cite{lattner2004llvm}. Any internal optimization modules must have maintained representations so that it is possible to add new optimizations~\footnote{Intermediate representations for visualization optimizations is not implemented yet.}.
        
        \textbf{Single Point of Access.} GAMMS centralizes all simulation state and control through a unified context object that serves as the single authoritative source for the current state of the simulation. This design is inspired from the very successful ``Cuda Context''~\cite{nvidiaCUDAProgramming} which hid the complexity from the user but has allowed development of LLMs with billions of parameters. GAMMS uses a context object that internally manages the implementations of the interfaces and allows the tool user to only think in terms of the small set of contracts defined by the maintained interfaces. Furthermore, a common access point prevents the problem of state inconsistency that arises when multiple components maintain their own copies of simulation data. By funneling all state access through a single point, the framework ensures that all components have a consistent view of the simulation state.

        In contrast to the strict and structured format of the design principles, we select Python as the language of choice for development. The primary reason is the low learning requirement for Python for the users, as well as the vast amount of libraries that have been developed over the years.

        Python is known to be a slow and heavy language, and many attempts have been made to improve the speed by creating separate implementations of the interpreter. Furthermore, Python is dynamically typed, meaning there is no enforced strict typing. While it may appear at first glance that the choice of Python is in complete contradiction, what we want to clarify is that the design principles themselves are completely orthogonal to the choice of any object-oriented programming language. The principles put restrictions solely on the developer and not on the user. As such, the developer can choose to use any language for implementation (even choose to write direct machine code) as long as the user can import the implementation as a normal Python package and get the behavior as described by the interface.

        Starting initially with Python has benefits for the development process as well. The first benefit is that the developers can balance between the time investment on new features and optimizations with ease. The second benefit is that it is possible to monkey-patch~\cite{Hunt2023} internally such that any call can be converted into a dispatch for a compiled implementation similar to PyTorch~\cite{pytorchRegisteringDispatched}. In general, there is always scope to internally optimize using just-in-time (JIT) compilation without the users having to make any changes (or minor changes) to their code.

    \subsection{Simulator Components}
    \label{subsec:simulator_components}
\begin{figure*}[ht]
            \centering
            \includegraphics[width=\linewidth]{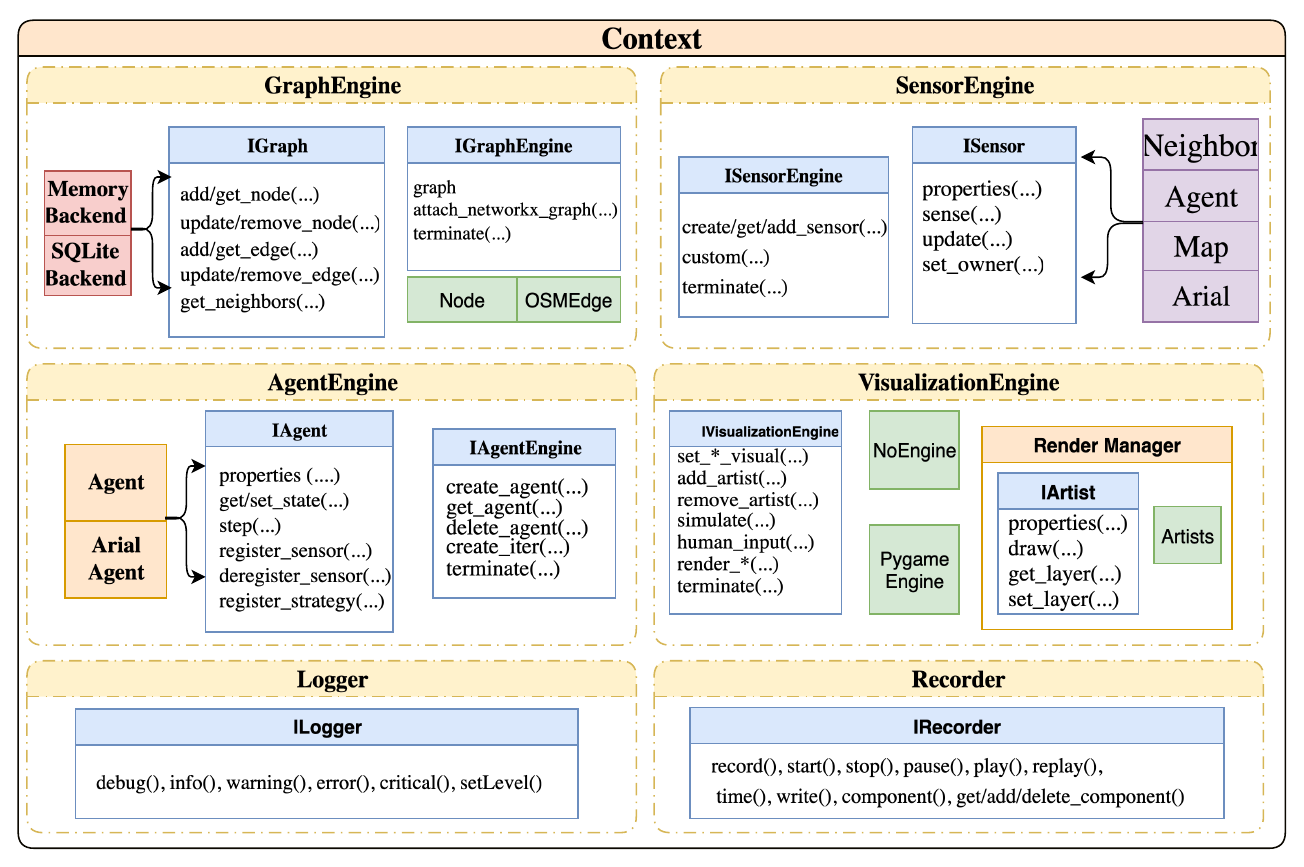}
            \caption{The Figure shows the current overall architecture of GAMMS.}
            \label{fig:architecture}
        \end{figure*}

        GAMMS splits sets of interfaces broadly into different parts such that each set is responsible for a specific task or tasks. This is primarily done so that it is easy to arrange the different interfaces in a hierarchical format. Figure \ref{fig:architecture} gives an overview of the internal structure of GAMMS.

        \textbf{Graph.} The Graph module in GAMMS is the component that defines the environment in which agents operate. It represents entities as nodes and their relationships as edges, which can be customized for different environments, such as networks or spatial layouts. The module is designed to be flexible, allowing it to model a range of structures, from simple connectivity graphs to more complex urban or communication networks. The Graph module consists of the \textit{Graph Engine} interface and the \textit{Graph} interface itself. The engine is supposed to be a gateway to create and update graphs, and maintain any form of context that may arise for the graph. The graph itself is a simple interface that allows adding, modifying, and deleting nodes and edges.

        \textbf{Sensor.} GAMMS treats a sensor as a module that is responsible for collecting data about the environment and the entire scenario. Every sensor needs to implement the \textit{sense} method, which is responsible for data collection. As the context object provides a single access point for the entire state of the scenario, it is easy to define custom sensors alongside a wide variety of sensors that are provided prepackaged. Furthermore, as the data flow does not only have to be from the context object, external information can also be injected into the agent through custom sensors. This provides the users flexibility to pass conditioned information that needs to be computed based on various factors. Alongside sensors, the module also consists of \textit{Sensor Engine} which is responsible for maintaining contexts and providing methods to ease the use of prepackaged sensors.
        \begin{minted}[
frame=lines,
fontsize=\footnotesize,
]{python}
ctx.sensor.create_sensor(
  'camera', type=gamms.sensor.SensorType.ARC, sensor_range=50, fov=2.5
) # Creates sensor named 'camera' that senses nodes and edges in range
ctx.visual.set_sensor_visual(
  'camera', node_color=(0, 255, 0), edge_color=(0, 255, 0)
) # Highlights the nodes and edges sensed by the sensor
        \end{minted}
        % \todo[inline]{Code for one sensor}

        \textbf{Agent.} The Agent module provides an API to register sensors, create the state by collecting data from registered sensors, and process the actions supplied. It is the container object responsible for maintaining the data and state associated with a particular agent. The separate implementation of agents provides a structure that allows it to be agnostic to the type of policy being used -- classical, deep learning method, or even human-controlled. Similar to previous modules, there is an \textit{Agent Engine} which maintains contexts and eases the interactions like adding or removing agents. The separation between agent logic and environment interaction also ensures that different agent behaviors can be easily swapped or modified without altering the simulation's structure.
        % \todo[inline]{Code for one agent}
        \begin{minted}[
frame=lines,
fontsize=\footnotesize,
]{python}
# Single line to create an agent
ctx.agent.create_agent("agent_0", start_node_id=0)
# Visualize agent_0 in red color on the graph
ctx.visual.set_agent_visual(name='agent_0',color=(255, 0, 0),size=10)
        \end{minted}

        \textbf{Recorder.} GAMMS provides a lightweight recorder which is integrated with other interface implementations that allow recording a scenario for replay. The recording process can be extremely unstructured, and hence it uses an \textit{Intermediate Representation} (IR) for recording the changes in the scenario. The recorder stores only agent interactions alongside custom user-defined data that needs to be recorded. With the assumption that the user ensures recording external data required for reproducing the entire scenario, it allows the recorder to create highly compressed recordings. Inspired by how StableHLO~\cite{openxlaStableHLOOpenXLA} provides compatibility between IRs across different versions, the recorder IR ensures that replaying an older version with a newer version of the recorder translates the older version recording to the newer version. Furthermore, there is an automatic check on correctness, similar to compilers~\cite{kumar2016self}~\footnote{A general check for compiler correctness is to recompile the source code of the compiler with itself to check if it produces the same binary file.}, as recording while replaying a recording from the same version should produce the exact same bytes.

        \textbf{Logger.} While setting up the context, it accepts arguments to get different levels of logs for debugging. This module is similar to any vanilla logger~\footnote{Current implementation uses the default python logger.}, but a strict interface is defined so that users can change it as required.

        \textbf{Visualization.} GAMMS employs a modular rendering pipeline built around a central Visualization Interface that abstracts the rendering process and delegates execution to specific backend implementations, such as a graphical PyGame engine or a headless NO\_VIS mode. When NO\_VIS is enabled, the rendering subsystem is bypassed entirely to eliminate graphical overhead, whereas enabling PYGAME activates full visual output. Visual elements in the simulation are represented as artists, each composed of primitive drawing operations like \emph{render\_rectangle} and \emph{render\_circle}, defined within the interface. During each visualization update, these abstract rendering calls are translated into backend-specific implementations, with optimizations such as view culling applied to skip rendering objects outside the visible window. Simultaneously, user input is processed within the same frame to ensure real-time responsiveness and maintain interactive performance. This roundabout way of handling visualization makes it easy to integrate different backends. It opens up the way to easily integrate something like Plotly~\cite{plotly}, potentially allowing support for web-based visualization.
        
        % \begin{figure}[t]
        %     \centering
        %     \includegraphics[width=.8\linewidth]{assets/VisGraph.png}
        %     \caption{A high level overview of the visualization system.}
        %     \label{fig:vis-graphic}
        % \end{figure}
        % \todo[inline]{Do we have an svg for the picture? If yes, can u convert it to a pdf and replace it? Also, in place of PygameEngine, can you make it Visualization interface? The arrow from render manager should be the other way ig. flow tldr -- ctx -> visualization interface -> render manager -> specific implementation (pygame engine or no vis engine) -> actually draw to window.}
        
\subsection{Overall Flow and Construction of the Game Loop}

% \todo[inline]{Add a overall figure showing flow of gamms ... similar to one on the docs website.}

\begin{figure*}[ht]
            \centering
            \includegraphics[width=\linewidth]{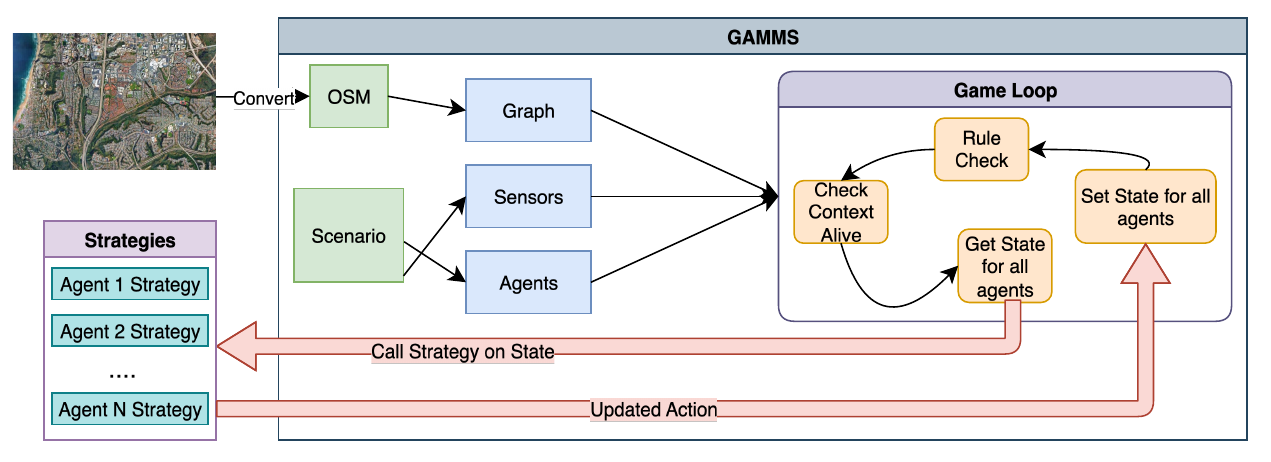}
            \caption{Overall flow and construction of a Game using GAMMS}
            \label{fig:flow}
        \end{figure*}

The game loop orchestrates the temporal progression of GAMMS simulations through a structured multi-phase approach. Each iteration consists of: environmental state synchronization, agent observation collection, strategy execution and action selection, custom rule enforcement, and visualization rendering. Figure \ref{fig:flow} shows the overall flow of a typical scenario from a user's perspective. 

\textbf{OSM Loading and Graph Construction.} GAMMS converts OpenStreetMap data into simulation-ready graph representations through a specialized processing pipeline that addresses key challenges in geographic data utilization for multi-agent simulation. Traditional approaches place nodes exclusively at intersections, creating irregular spacing that complicates agent movement and sensor modeling. GAMMS addresses this through uniform spatial resampling, subdividing road segments into nodes spaced at consistent intervals. The processing pipeline uses OSMnx~\cite{Boeing2025} to do edge resampling to ensure uniform node spacing, intersection consolidation to merge nearby nodes representing the same physical intersection, and bidirectional edge handling that respects transportation mode constraints while providing experimental flexibility.

\begin{minted}[
frame=lines,
fontsize=\footnotesize,
]{python}
# Location-based loading with automatic processing
G = gamms.osm.create_osm_graph(
    "La Jolla, San Diego, CA, USA", resolution=10)
# Or .. XML-based loading for custom scenarios
G = gamms.osm.graph_from_xml("manhattan_map.osm", resolution=5)
ctx.graph.attach_networkx_graph(G)
\end{minted}

\textbf{User Strategies and Policy Architecture.} GAMMS employs a compositional strategy pattern where agent behaviors are implemented as callable functions that accept and modify agent state. This functional approach enables strategy composition and runtime behavior modification without simulation restart.

\begin{minted}[
frame=lines,
framesep=2mm,
baselinestretch=1.2,
fontsize=\footnotesize,
]{python}
# red_strategy.py - Modular strategy implementation
def strategy(state):
    sensor_data = state['sensor']
    for sensor_type, data in sensor_data.values():
        if sensor_type == sensor.SensorType.NEIGHBOR:
            choice = random.choice(range(len(data)))
            state['action'] = data[choice]
            break
def map_strategy(agent_config):
    strategies = {}
    for name in agent_config.keys():
        strategies[name] = strategy
    return strategies
\end{minted}
Strategy registration supports both individual agent customization and team-based behavior patterns, enabling heterogeneous agent populations within single simulations.

\begin{minted}[
frame=lines,
framesep=2mm,
baselinestretch=1.2,
fontsize=\footnotesize,
]{python}
# Team-based strategy assignment
red_agents = {name: config for name, config in agent_config.items() 
              if config['meta']['team'] == 1}
strategies = red_strategy.map_strategy(red_agents)
for agent in ctx.agent.create_iter():
    agent_strategy = strategies.get(agent.name, None)
    agent.register_strategy(agent_strategy)
\end{minted}

As the entire strategy implementation is only linked via a single function call, the strategy is completely isolated from the simulator. Having such a strategy agnostic structure gives the user the flexibility to work with any classical rule-based strategies, machine learning policies, and human-controlled agents within the same simulation environment. Human players interact through the visualization interface, enabling human-in-the-loop experiments and direct comparison of human versus algorithmic decision-making under identical conditions.

% Agent state includes current position and node information, sensor data from registered sensors, team membership, and custom variables defined by the simulation scenario. This comprehensive state representation enables sophisticated behaviors including coordination, territorial control, and long-term planning.

\textbf{Game Rules Execution and Simulation Loop.} The execution framework implements a multi-phase game loop that maintains state consistency across large agent populations while resolving action conflicts deterministically.

% \begin{minted}[
% frame=lines,
% framesep=2mm,
% baselinestretch=1.2,
% fontsize=\footnotesize,
% ]{python}
% # Comprehensive simulation initialization
% ctx = gamms.create_context()

% # Sensor system setup with validation
% for sensor_name, config in sensor_config.items():
%     ctx.sensor.create_sensor(sensor_name, config['type'], **config)

% # Agent initialization with sensor registration
% for agent_name, config in agent_config.items():
%     agent = ctx.agent.create_agent(agent_name, start_node=config['start_node'])
%     for sensor_name in config['sensors']:
%         sensor_obj = ctx.sensor.get_sensor(sensor_name)
%         agent.register_sensor(sensor_name, sensor_obj)
    
%     # Strategy registration
%     strategy = strategies.get(agent_name, None)
%     agent.register_strategy(strategy)
% \end{minted}

The core simulation loop processes all agents simultaneously within each phase, preventing timing dependencies and ensuring fair competition between different strategies.

\begin{minted}[
frame=lines,
framesep=2mm,
baselinestretch=1.2,
fontsize=\footnotesize,
]{python}
turn_count = 0
while not ctx.is_terminated():
    turn_count += 1
    # Agent observation and decision phase
    for agent in ctx.agent.create_iter():
        state = agent.get_state()  # Collect sensor data Dict[str, Any]
        agent.strategy(state) # Updates state with 'action' entry
        agent.set_state() # Processes the updated 'action' entry
    rule1(ctx, turn_count) # Call the rule which is just a function    
    ctx.visual.simulate() # Visualization and recording
\end{minted}

The action conflict resolution system handles situations where multiple agents attempt conflicting actions through configurable resolution mechanisms including priority-based systems, random selection, and custom logic tailored to specific experimental requirements.

The custom rule system enables domain-specific simulation logic while maintaining separation between the simulation engine and experimental requirements. Rules execute after agent actions but before visualization, providing the ability to modify simulation state, enforce constraints, implement scoring systems, and determine termination conditions. Rules can implement mechanics ranging from simple win/loss conditions to complex resource management, territory control, or communication protocols between agents.

The visualization and recording integration provides real-time observation capabilities for development and analysis, while comprehensive data capture supports offline analysis and reproducible research practices.

\section{Examples}
\label{sec:examples}
    In this section, we will explore how GAMMS adapts to both simple scenarios, such as a grid world, and more complex ones, like a capture-the-flag setup, while also supporting both ground and aerial agents.

\begin{figure}[h]
            \centering
            \begin{minipage}[c]{0.5\linewidth}
                \includegraphics[width=\linewidth]{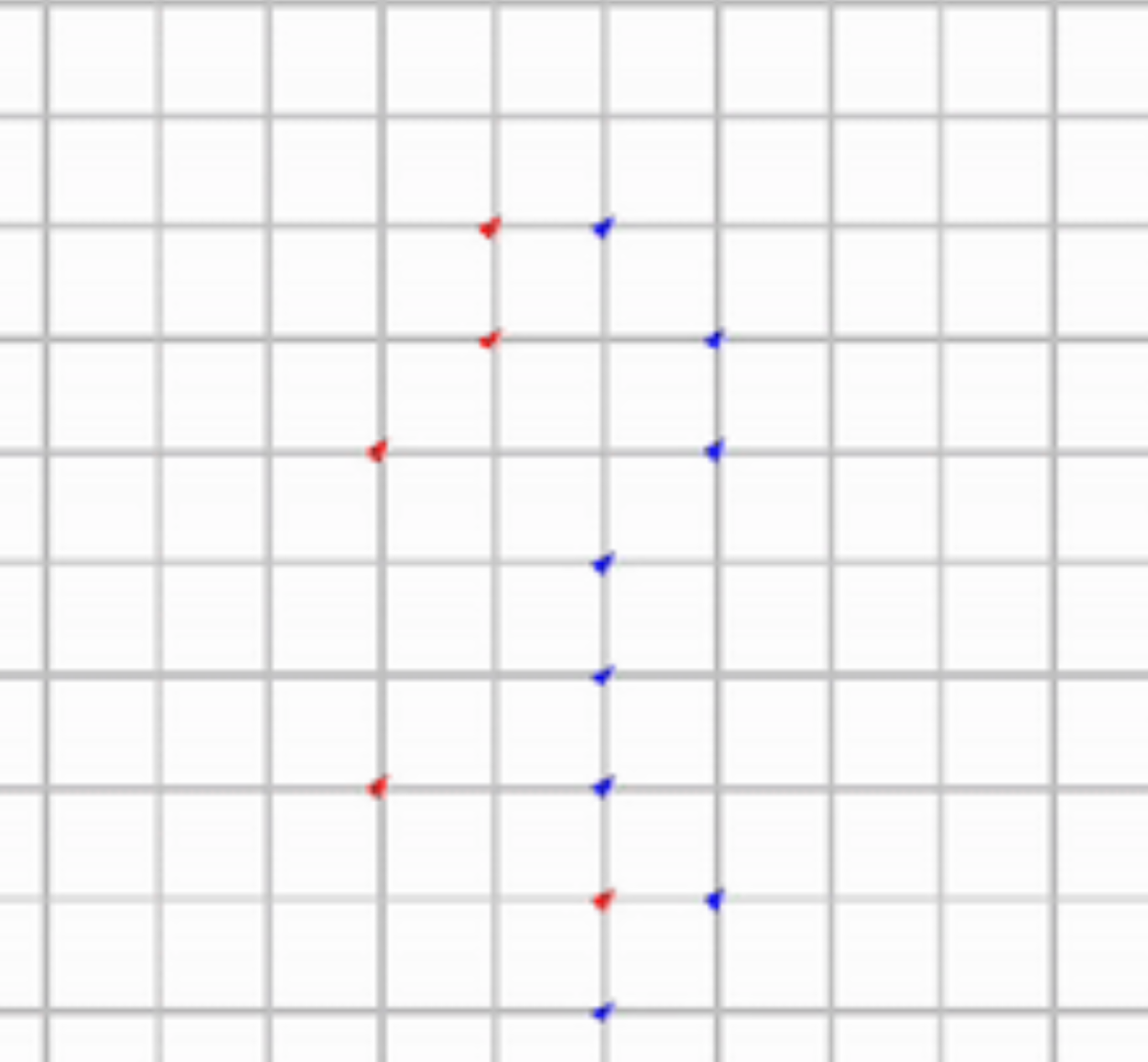}                
            \end{minipage}
            \begin{minipage}[c]{0.4\linewidth}
                \caption{Part of the grid world created by GAMMS.}
                \label{fig:grid_world}
            \end{minipage}
        \end{figure}

    \textbf{Grid World.} Grid World is a discrete multi-agent environment where agents move on a lattice of nodes. It serves as a simple testbed for exploring coordination, path-finding, and basic strategies before scaling to real-world scenarios. A grid can be created manually by adding nodes and edges to a graph object. Suppose we want to have two teams, the blue team and the red team, where the objective of the teams is to reach the starting positions of the other team first. We can easily create some names like \emph{red\_1}, \emph{blue\_1} for the agents and store them. Figure \ref{fig:grid_world} shows the visualization created by GAMMS.
                
        \begin{minted}[
        frame=lines,
        framesep=2mm,
        baselinestretch=1.2,
        fontsize=\footnotesize,
        ]{python}
def create_grid(graph, n):
  count = 0 # initialize the edge count to 0
  for i in range(n):
    for j in range(n):
      ... # Use add_node and add_edge
red_team = [...]; blue_team = [...]
def tag_rule(ctx):
  for red in red_team:
    for blue in blue_team:
      ragent = ctx.agent.get_agent(red)
      bagent = ctx.agent.get_agent(blue)
      if ragent.current_node_id == bagent.current_node_id:
        ... # Reset to starting positions
create_grid(ctx.graph.graph, 20) # Generate a 20x20 grid
graph_artist = ctx.visual.set_graph_visual(width=400, height=400)
# ... create agents and set color according to team
while not ctx.is_terminated(): # run loop until context is terminated
  # ... iterate over agents
  tag_rule(ctx) # Check for taggings
        \end{minted}

\begin{figure}[h]
            \centering
            \begin{minipage}[c]{0.5\linewidth}
                \includegraphics[width=\linewidth]{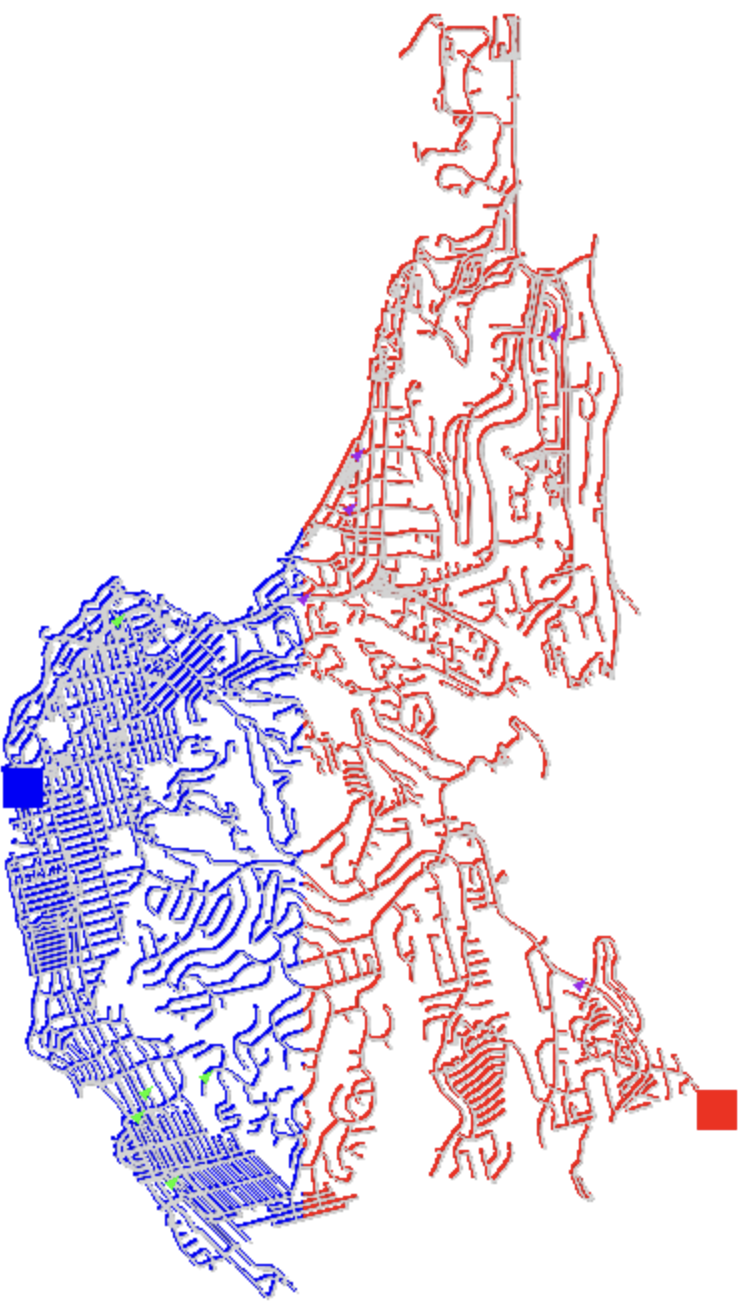}                
            \end{minipage}
            \begin{minipage}[c]{0.4\linewidth}
                \caption{Part of La Jolla OSM converted into a GAMMS graph for the Capture the Flag scenario. Red and blue areas show the territories of the respective teams. The big sqares show the flag positions.}
                \label{fig:ctf}
            \end{minipage}
        \end{figure}

\textbf{Capture the Flag.} The base grid world can easily be made more realistic by first swapping the graph with a real-world map. Figure \ref{fig:ctf} shows a part of the La Jolla area in California. Here, the two teams have their own territories highlighted in the figure by their team color. The tagging mechanic is changed such that only the opponent agent with respect to the territory gets reset. Both teams want to capture the flag of the other team first. The entire figure was made purely using GAMMS via its custom drawing API. The \emph{artist} API allows the user to create overlays directly in the world frame.

        \begin{minted}[
        frame=lines,
        baselinestretch=1.2,
        fontsize=\footnotesize,
        ]{python}
def draw_flags(ctx, data): # Example artist for drawing flags
  red_id = data.get('red_id') # Node id for red flag
  blue_id = data.get('blue_id') # Node id for blue flag
  blue = ctx.graph.graph.get_node(blue_id)
  red = ctx.graph.graph.get_node(red_id)
  ctx.visual.render_rectangle(blue.x, blue.y, 10, 10, color=(0,255,0))
  ctx.visual.render_rectangle(red.x, red.y, 10, 10, color=(255,0,0))
# Create artist with layer above the graph
capturable_artist = gamms.visual.Artist(ctx,drawer=draw_flags,layer=35)
capturable_artist.data['red_id] = ...
capturable_artist.data['blue_id] = ...
ctx.visual.add_artist('capturable_artist', capturable_artist)
        \end{minted}
        
        % \todo[inline]{ Capture the Flag scenario with two teams, flags, and home bases.}

\begin{figure}[h]
            \centering
            \includegraphics[width=\linewidth]{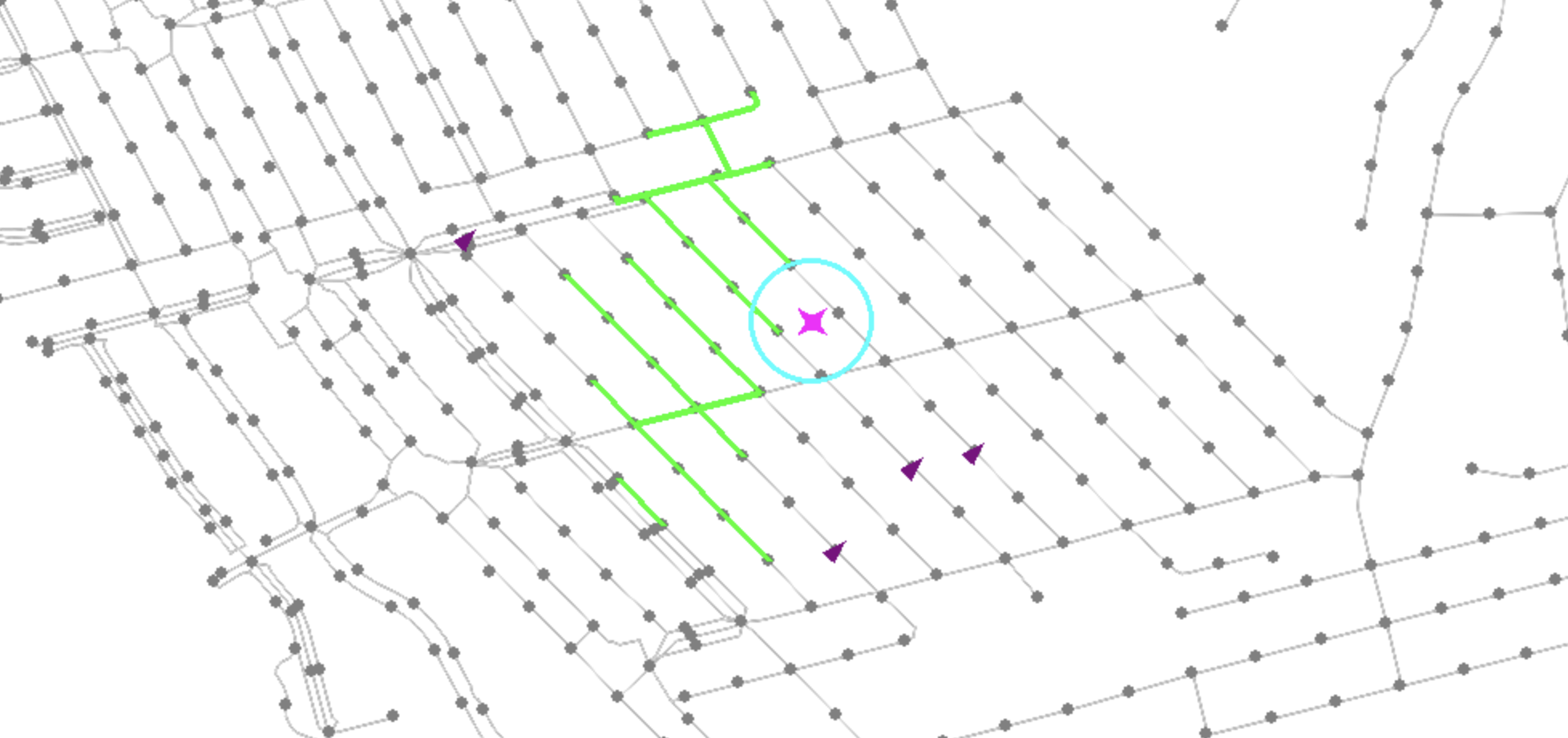}
            \caption{Figure shows a scenario with aerial agents and ground agents. The highlighted nodes and edges are the ones that are visible to the camera of the aerial agent.}
            \label{fig:aerial}
        \end{figure}

    \textbf{Heterogeneous Robot Systems.} A real-world multi-agent system usually has different types of agents that complement each other's abilities to achieve tasks. Drones or aerial agents are great for collecting environmental information, but have limitations with direct interaction. Furthermore, in scenarios like fires or areas with many humans, flying a drone closer to the ground can be a safety risk. Hence, we would ideally want a tag-team of aerial agents collecting information for ground agents. GAMMS supports aerial agents in addition to ground agents so that it is possible to test such complex team dynamics. Figure \ref{fig:aerial} shows an aerial agent along with some other ground agents.

    \begin{minted}[
        frame=lines,
        baselinestretch=1.2,
        fontsize=\footnotesize,
        ]{python}
ctx.agent.create_agent(
  'aerial', agent_type=gamms.agent.AgentType.AERIAL,
  start_node_id=0, speed=1.0
) # Creates an aerial agent that starts on the ground       
    \end{minted}

\section{Discussion \& Future Work}
\label{sec:discussion}

    % Create citation: https://www.cudahandbook.com/2017/06/ten-years-later-why-cuda-succeeded
    Simulator design is not just about efficient implementations but more about design choices that put the users first. Gymnasium~\cite{towers2024gymnasium}, which is a popular choice for defining RL environments, saw its success because of good design choices and provided a structure for researchers and developers to work with. Similarly, the success of CUDA was because of its ease of use~\cite{lattnerhow2025}. What sets GAMMS aside is that each design choice involved was based on the successes and failures of various frameworks, cherry-picking the choices that made the individual tools favorable to the users. At present, it is being used across six universities and one U.S. government research lab.

    Usually when starting with simple examples, it becomes easier for users, particularly researchers, to have a clean single-file, high-quality implementation. From the various code examples we saw in previous sections, it is clear that a user can start working on small examples quickly by writing a small file containing a few hundred lines to get a fully functioning scenario with some basic visualization. The ability to get quick feedback and iterate allows quick development of complex scenarios. Furthermore, the entire package can be installed using any python package manager like \emph{pip}. We intend to keep it as a hard constraint for any future development as the users should never have to wrestle with the toolchain just to get started.

    Planned future work includes refining the current feature set and adding more integrated examples, especially those that show how to use external libraries for agent development. There is also ongoing work on transferring policies trained in GAMMS to real robots at WestPoint. GAMMS remains a work in progress, but its current design and usage indicate that it provides a practical starting point for building and experimenting with multi-agent systems.

%%%%%%%%%%%%%%%%%%%%%%%%%%%%%%%%%%%%%%%%%%%%%%%%%%%%%%%%%%%%%%%%%%%%%%%%

%%% The acknowledgments section is defined using the "acks" environment
%%% (rather than an unnumbered section). The use of this environment 
%%% ensures the proper identification of the section in the article 
%%% metadata as well as the consistent spelling of the heading.

\begin{acks}
We gratefully acknowledge support from ARL DCIST CRA W911NF-17-2-0181
\end{acks}

%%%%%%%%%%%%%%%%%%%%%%%%%%%%%%%%%%%%%%%%%%%%%%%%%%%%%%%%%%%%%%%%%%%%%%%%

%%% The next two lines define, first, the bibliography style to be 
%%% applied, and, second, the bibliography file to be used.

\bibliographystyle{ACM-Reference-Format} 
\bibliography{ref}


\begin{thebibliography}{32}

%%% the Web address.

\ifx \showCODEN    \undefined \def \showCODEN     #1{\unskip}     \fi
\ifx \showDOI      \undefined \def \showDOI       #1{#1}\fi
\ifx \showISBNx    \undefined \def \showISBNx     #1{\unskip}     \fi
\ifx \showISBNxiii \undefined \def \showISBNxiii  #1{\unskip}     \fi
\ifx \showISSN     \undefined \def \showISSN      #1{\unskip}     \fi
\ifx \showLCCN     \undefined \def \showLCCN      #1{\unskip}     \fi
\ifx \shownote     \undefined \def \shownote      #1{#1}          \fi
\ifx \showarticletitle \undefined \def \showarticletitle #1{#1}   \fi
\ifx \showURL      \undefined \def \showURL       {\relax}        \fi
% invisible to TeX
\providecommand\bibfield[2]{#2}
\providecommand\bibinfo[2]{#2}
\providecommand\natexlab[1]{#1}
\providecommand\showeprint[2][]{arXiv:#2}

\bibitem[\protect\citeauthoryear{??}{ope}{[n.d.]}]%
        {openxlaStableHLOOpenXLA}
 \bibinfo{year}{[n.d.]}\natexlab{}.
\newblock \bibinfo{title}{{S}table{H}{L}{O} | {O}pen{X}{L}{A} {P}roject ---
  openxla.org}.
\newblock \bibinfo{howpublished}{\url{https://openxla.org/stablehlo}}.
\newblock
\newblock
\shownote{[Accessed 22-09-2025].}


\bibitem[\protect\citeauthoryear{??}{pyt}{2020}]%
        {pytorchRegisteringDispatched}
 \bibinfo{year}{2020}\natexlab{}.
\newblock \bibinfo{title}{{R}egistering a {D}ispatched {O}perator in {C}++
  \#x2014; {P}y{T}orch {T}utorials 2.8.0+cu128 documentation ---
  docs.pytorch.org}.
\newblock
  \bibinfo{howpublished}{\url{https://docs.pytorch.org/tutorials/advanced/dispatcher}}.
\newblock
\newblock
\shownote{[Accessed 22-09-2025].}


\bibitem[\protect\citeauthoryear{Amouroux, Chu, Boucher, and Drogoul}{Amouroux
  et~al\mbox{.}}{2009}]%
        {Amouroux2009}
\bibfield{author}{\bibinfo{person}{Edouard Amouroux},
  \bibinfo{person}{Thanh-Quang Chu}, \bibinfo{person}{Alain Boucher}, {and}
  \bibinfo{person}{Alexis Drogoul}.} \bibinfo{year}{2009}\natexlab{}.
\newblock \bibinfo{booktitle}{\emph{GAMA: An Environment for Implementing and
  Running Spatially Explicit Multi-agent Simulations}}.
\newblock \bibinfo{publisher}{Springer Berlin Heidelberg},
  \bibinfo{pages}{359–371}.
\newblock
\showISBNx{9783642016394}
\showISSN{1611-3349}
\urldef\tempurl%
\url{https://doi.org/10.1007/978-3-642-01639-4_32}
\showDOI{\tempurl}


\bibitem[\protect\citeauthoryear{Bahrin, Othman, Azli, and Talib}{Bahrin
  et~al\mbox{.}}{2016}]%
        {bahrin2016industry}
\bibfield{author}{\bibinfo{person}{Mohd Aiman~Kamarul Bahrin},
  \bibinfo{person}{Mohd~Fauzi Othman}, \bibinfo{person}{Nor Hayati~Nor Azli},
  {and} \bibinfo{person}{Muhamad~Farihin Talib}.}
  \bibinfo{year}{2016}\natexlab{}.
\newblock \showarticletitle{Industry 4.0: A review on industrial automation and
  robotic}.
\newblock \bibinfo{journal}{\emph{Jurnal Teknologi (Sciences \& Engineering)}}
  \bibinfo{volume}{78}, \bibinfo{number}{6-13} (\bibinfo{year}{2016}).
\newblock


\bibitem[\protect\citeauthoryear{Beyer and Kanav}{Beyer and Kanav}{2020}]%
        {beyer2020interface}
\bibfield{author}{\bibinfo{person}{Dirk Beyer} {and} \bibinfo{person}{Sudeep
  Kanav}.} \bibinfo{year}{2020}\natexlab{}.
\newblock \showarticletitle{An interface theory for program verification}. In
  \bibinfo{booktitle}{\emph{International Symposium on Leveraging Applications
  of Formal Methods}}. Springer, \bibinfo{pages}{168--186}.
\newblock


\bibitem[\protect\citeauthoryear{Boeing}{Boeing}{2025}]%
        {Boeing2025}
\bibfield{author}{\bibinfo{person}{Geoff Boeing}.}
  \bibinfo{year}{2025}\natexlab{}.
\newblock \showarticletitle{Modeling and Analyzing Urban Networks and Amenities
  With OSMnx}.
\newblock \bibinfo{journal}{\emph{Geographical Analysis}} (\bibinfo{date}{May}
  \bibinfo{year}{2025}).
\newblock
\showISSN{1538-4632}
\urldef\tempurl%
\url{https://doi.org/10.1111/gean.70009}
\showDOI{\tempurl}


\bibitem[\protect\citeauthoryear{Chadha and Docca}{Chadha and Docca}{2024}]%
        {rishabh2024beginners}
\bibfield{author}{\bibinfo{person}{Rishabh Chadha} {and} \bibinfo{person}{Akhil
  Docca}.} \bibinfo{year}{2024}\natexlab{}.
\newblock \bibinfo{title}{A Beginner’s Guide to Simulating and Testing Robots
  with ROS 2 and NVIDIA Isaac Sim}.
\newblock
  \bibinfo{howpublished}{\url{https://developer.nvidia.com/blog/a-beginners-guide-to-simulating-and-testing-robots-with-ros-2-and-nvidia-isaac-sim/}}.
\newblock
\newblock
\shownote{[Accessed 26-09-2025].}


\bibitem[\protect\citeauthoryear{Datseris, Vahdati, and DuBois}{Datseris
  et~al\mbox{.}}{2022}]%
        {Agents.jl}
\bibfield{author}{\bibinfo{person}{George Datseris}, \bibinfo{person}{Ali~R.
  Vahdati}, {and} \bibinfo{person}{Timothy~C. DuBois}.}
  \bibinfo{year}{2022}\natexlab{}.
\newblock \showarticletitle{Agents.jl: a performant and feature-full
  agent-based modeling software of minimal code complexity}.
\newblock \bibinfo{journal}{\emph{{SIMULATION}}} \bibinfo{volume}{0},
  \bibinfo{number}{0} (\bibinfo{date}{Jan.} \bibinfo{year}{2022}),
  \bibinfo{pages}{003754972110688}.
\newblock
\urldef\tempurl%
\url{https://doi.org/10.1177/00375497211068820}
\showDOI{\tempurl}


\bibitem[\protect\citeauthoryear{Forlizzi and DiSalvo}{Forlizzi and
  DiSalvo}{2006}]%
        {forlizzi2006service}
\bibfield{author}{\bibinfo{person}{Jodi Forlizzi} {and} \bibinfo{person}{Carl
  DiSalvo}.} \bibinfo{year}{2006}\natexlab{}.
\newblock \showarticletitle{Service robots in the domestic environment: a study
  of the roomba vacuum in the home}. In \bibinfo{booktitle}{\emph{Proceedings
  of the 1st ACM SIGCHI/SIGART conference on Human-robot interaction}}.
  \bibinfo{pages}{258--265}.
\newblock


\bibitem[\protect\citeauthoryear{Hunt}{Hunt}{2023}]%
        {Hunt2023}
\bibfield{author}{\bibinfo{person}{John Hunt}.}
  \bibinfo{year}{2023}\natexlab{}.
\newblock \bibinfo{booktitle}{\emph{Monkey Patching}}.
\newblock \bibinfo{publisher}{Springer International Publishing},
  \bibinfo{address}{Cham}, \bibinfo{pages}{487--490}.
\newblock
\showISBNx{978-3-031-35122-8}
\urldef\tempurl%
\url{https://doi.org/10.1007/978-3-031-35122-8_43}
\showDOI{\tempurl}


\bibitem[\protect\citeauthoryear{Iio, Satake, Kanda, Hayashi, Ferreri, and
  Hagita}{Iio et~al\mbox{.}}{2020}]%
        {iio2020human}
\bibfield{author}{\bibinfo{person}{Takamasa Iio}, \bibinfo{person}{Satoru
  Satake}, \bibinfo{person}{Takayuki Kanda}, \bibinfo{person}{Kotaro Hayashi},
  \bibinfo{person}{Florent Ferreri}, {and} \bibinfo{person}{Norihiro Hagita}.}
  \bibinfo{year}{2020}\natexlab{}.
\newblock \showarticletitle{Human-like guide robot that proactively explains
  exhibits}.
\newblock \bibinfo{journal}{\emph{International Journal of Social Robotics}}
  \bibinfo{volume}{12}, \bibinfo{number}{2} (\bibinfo{year}{2020}),
  \bibinfo{pages}{549--566}.
\newblock


\bibitem[\protect\citeauthoryear{Inc.}{Inc.}{2015}]%
        {plotly}
\bibfield{author}{\bibinfo{person}{Plotly~Technologies Inc.}}
  \bibinfo{year}{2015}\natexlab{}.
\newblock \bibinfo{booktitle}{\emph{Collaborative data science}}.
\newblock Montreal, QC.
\newblock
\urldef\tempurl%
\url{https://plot.ly}
\showURL{%
\tempurl}


\bibitem[\protect\citeauthoryear{Kim, Kim, Choi, Park, Oh, and Park}{Kim
  et~al\mbox{.}}{2024}]%
        {kim2024survey}
\bibfield{author}{\bibinfo{person}{Yeseung Kim}, \bibinfo{person}{Dohyun Kim},
  \bibinfo{person}{Jieun Choi}, \bibinfo{person}{Jisang Park},
  \bibinfo{person}{Nayoung Oh}, {and} \bibinfo{person}{Daehyung Park}.}
  \bibinfo{year}{2024}\natexlab{}.
\newblock \showarticletitle{A survey on integration of large language models
  with intelligent robots}.
\newblock \bibinfo{journal}{\emph{Intelligent Service Robotics}}
  \bibinfo{volume}{17}, \bibinfo{number}{5} (\bibinfo{year}{2024}),
  \bibinfo{pages}{1091--1107}.
\newblock


\bibitem[\protect\citeauthoryear{Kumar}{Kumar}{2016}]%
        {kumar2016self}
\bibfield{author}{\bibinfo{person}{Ramana Kumar}.}
  \bibinfo{year}{2016}\natexlab{}.
\newblock \bibinfo{booktitle}{\emph{Self-compilation and self-verification}}.
\newblock \bibinfo{type}{{T}echnical {R}eport}.
  \bibinfo{institution}{University of Cambridge, Computer Laboratory}.
\newblock


\bibitem[\protect\citeauthoryear{Labiosa, Wang, Agarwal, Cong, Hemkumar,
  Harish, Hong, Kelle, Li, Li, et~al\mbox{.}}{Labiosa et~al\mbox{.}}{2025}]%
        {labiosa2025reinforcement}
\bibfield{author}{\bibinfo{person}{Adam Labiosa}, \bibinfo{person}{Zhihan
  Wang}, \bibinfo{person}{Siddhant Agarwal}, \bibinfo{person}{William Cong},
  \bibinfo{person}{Geethika Hemkumar}, \bibinfo{person}{Abhinav~Narayan
  Harish}, \bibinfo{person}{Benjamin Hong}, \bibinfo{person}{Josh Kelle},
  \bibinfo{person}{Chen Li}, \bibinfo{person}{Yuhao Li}, {et~al\mbox{.}}}
  \bibinfo{year}{2025}\natexlab{}.
\newblock \showarticletitle{Reinforcement learning within the classical
  robotics stack: A case study in robot soccer}. In
  \bibinfo{booktitle}{\emph{2025 IEEE International Conference on Robotics and
  Automation (ICRA)}}. IEEE, \bibinfo{pages}{14999--15006}.
\newblock


\bibitem[\protect\citeauthoryear{Lattner}{Lattner}{2025}]%
        {lattnerhow2025}
\bibfield{author}{\bibinfo{person}{Chris Lattner}.}
  \bibinfo{year}{2025}\natexlab{}.
\newblock \bibinfo{title}{{Modular: How did CUDA succeed? (Democratizing AI
  Compute, Part 3)}}.
\newblock
\newblock
\urldef\tempurl%
\url{https://www.modular.com/blog/democratizing-ai-compute-part-3-how-did-cuda-succeed\#:~:text=The%20benefits%20of%20this%20approach%20were%20twofold:,created%2C%20making%20the%20platform%20even%20more%20valuable.}
\showURL{%
\tempurl}


\bibitem[\protect\citeauthoryear{Lattner and Adve}{Lattner and Adve}{2004}]%
        {lattner2004llvm}
\bibfield{author}{\bibinfo{person}{Chris Lattner} {and} \bibinfo{person}{Vikram
  Adve}.} \bibinfo{year}{2004}\natexlab{}.
\newblock \showarticletitle{LLVM: A compilation framework for lifelong program
  analysis \& transformation}. In \bibinfo{booktitle}{\emph{International
  symposium on code generation and optimization, 2004. CGO 2004.}} IEEE,
  \bibinfo{pages}{75--86}.
\newblock


\bibitem[\protect\citeauthoryear{Liang, Makoviychuk, Handa, Chentanez, Macklin,
  and Fox}{Liang et~al\mbox{.}}{2018}]%
        {liang2018gpu}
\bibfield{author}{\bibinfo{person}{Jacky Liang}, \bibinfo{person}{Viktor
  Makoviychuk}, \bibinfo{person}{Ankur Handa}, \bibinfo{person}{Nuttapong
  Chentanez}, \bibinfo{person}{Miles Macklin}, {and} \bibinfo{person}{Dieter
  Fox}.} \bibinfo{year}{2018}\natexlab{}.
\newblock \showarticletitle{Gpu-accelerated robotic simulation for distributed
  reinforcement learning}. In \bibinfo{booktitle}{\emph{Conference on Robot
  Learning}}. PMLR, \bibinfo{pages}{270--282}.
\newblock


\bibitem[\protect\citeauthoryear{Luke, Cioffi-Revilla, Panait, and
  Sullivan}{Luke et~al\mbox{.}}{2005}]%
        {Luke2005MASON}
\bibfield{author}{\bibinfo{person}{Sean Luke}, \bibinfo{person}{Claudio
  Cioffi-Revilla}, \bibinfo{person}{Liviu Panait}, {and} \bibinfo{person}{Keith
  Sullivan}.} \bibinfo{year}{2005}\natexlab{}.
\newblock \showarticletitle{MASON: A Multiagent Simulation Environment}.
\newblock \bibinfo{journal}{\emph{SIMULATION}} \bibinfo{volume}{81},
  \bibinfo{number}{7} (\bibinfo{year}{2005}), \bibinfo{pages}{517--527}.
\newblock
\urldef\tempurl%
\url{https://doi.org/10.1177/0037549705058073}
\showDOI{\tempurl}


\bibitem[\protect\citeauthoryear{Macenski, Foote, Gerkey, Lalancette, and
  Woodall}{Macenski et~al\mbox{.}}{2022}]%
        {doi:10.1126/scirobotics.abm6074}
\bibfield{author}{\bibinfo{person}{Steven Macenski}, \bibinfo{person}{Tully
  Foote}, \bibinfo{person}{Brian Gerkey}, \bibinfo{person}{Chris Lalancette},
  {and} \bibinfo{person}{William Woodall}.} \bibinfo{year}{2022}\natexlab{}.
\newblock \showarticletitle{Robot Operating System 2: Design, architecture, and
  uses in the wild}.
\newblock \bibinfo{journal}{\emph{Science Robotics}} \bibinfo{volume}{7},
  \bibinfo{number}{66} (\bibinfo{year}{2022}), \bibinfo{pages}{eabm6074}.
\newblock
\urldef\tempurl%
\url{https://doi.org/10.1126/scirobotics.abm6074}
\showDOI{\tempurl}


\bibitem[\protect\citeauthoryear{Meyer}{Meyer}{2002}]%
        {meyer2002applying}
\bibfield{author}{\bibinfo{person}{Bertrand Meyer}.}
  \bibinfo{year}{2002}\natexlab{}.
\newblock \showarticletitle{Applying'design by contract'}.
\newblock \bibinfo{journal}{\emph{Computer}} \bibinfo{volume}{25},
  \bibinfo{number}{10} (\bibinfo{year}{2002}), \bibinfo{pages}{40--51}.
\newblock


\bibitem[\protect\citeauthoryear{Noorani, Serlin, Price, and Velasquez}{Noorani
  et~al\mbox{.}}{2025}]%
        {noorani2025abstraction}
\bibfield{author}{\bibinfo{person}{Erfaun Noorani}, \bibinfo{person}{Zachary
  Serlin}, \bibinfo{person}{Ben Price}, {and} \bibinfo{person}{Alvaro
  Velasquez}.} \bibinfo{year}{2025}\natexlab{}.
\newblock \showarticletitle{From abstraction to reality: DARPA's vision for
  robust sim-to-real autonomy}.
\newblock \bibinfo{journal}{\emph{AI Magazine}} \bibinfo{volume}{46},
  \bibinfo{number}{2} (\bibinfo{year}{2025}), \bibinfo{pages}{e70015}.
\newblock


\bibitem[\protect\citeauthoryear{North, Collier, Ozik, Tatara, Macal, Bragen,
  and Sydelko}{North et~al\mbox{.}}{2013}]%
        {North2013}
\bibfield{author}{\bibinfo{person}{Michael~J North},
  \bibinfo{person}{Nicholson~T Collier}, \bibinfo{person}{Jonathan Ozik},
  \bibinfo{person}{Eric~R Tatara}, \bibinfo{person}{Charles~M Macal},
  \bibinfo{person}{Mark Bragen}, {and} \bibinfo{person}{Pam Sydelko}.}
  \bibinfo{year}{2013}\natexlab{}.
\newblock \showarticletitle{Complex adaptive systems modeling with Repast
  Simphony}.
\newblock \bibinfo{journal}{\emph{Complex Adaptive Systems Modeling}}
  \bibinfo{volume}{1}, \bibinfo{number}{1} (\bibinfo{date}{March}
  \bibinfo{year}{2013}).
\newblock
\showISSN{2194-3206}
\urldef\tempurl%
\url{https://doi.org/10.1186/2194-3206-1-3}
\showDOI{\tempurl}


\bibitem[\protect\citeauthoryear{NVIDIA}{NVIDIA}{2014}]%
        {nvidiaCUDAProgramming}
\bibfield{author}{\bibinfo{person}{NVIDIA}.} \bibinfo{year}{2014}\natexlab{}.
\newblock \bibinfo{title}{{C}{U}{D}{A} {C}++ {P}rogramming {G}uide x2014;
  {C}{U}{D}{A} {C}++ {P}rogramming {G}uide --- docs.nvidia.com}.
\newblock
  \bibinfo{howpublished}{\url{https://docs.nvidia.com/cuda/cuda-c-programming-guide/index.html?highlight=context\#initialization}}.
\newblock
\newblock
\shownote{[Accessed 22-09-2025].}


\bibitem[\protect\citeauthoryear{of~Transport}{of~Transport}{2022}]%
        {nycInfrastructureTraffic}
\bibfield{author}{\bibinfo{person}{New York City~Department of Transport}.}
  \bibinfo{year}{2022}\natexlab{}.
\newblock \bibinfo{title}{{N}{Y}{C} {D}{O}{T} - {I}nfrastructure - {T}raffic
  {S}ignals --- nyc.gov}.
\newblock
  \bibinfo{howpublished}{\url{https://www.nyc.gov/html/dot/html/infrastructure/signals.shtml}}.
\newblock
\newblock
\shownote{[Accessed 22-09-2025].}


\bibitem[\protect\citeauthoryear{{OpenStreetMap contributors}}{{OpenStreetMap
  contributors}}{2017}]%
        {OpenStreetMap}
\bibfield{author}{\bibinfo{person}{{OpenStreetMap contributors}}.}
  \bibinfo{year}{2017}\natexlab{}.
\newblock \bibinfo{title}{{Planet dump retrieved from https://planet.osm.org
  }}.
\newblock \bibinfo{howpublished}{\url{ https://www.openstreetmap.org }}.
\newblock


\bibitem[\protect\citeauthoryear{Sung and Jeon}{Sung and Jeon}{2020}]%
        {sung2020untact}
\bibfield{author}{\bibinfo{person}{Hye~Jin Sung} {and}
  \bibinfo{person}{Hyeon~Mo Jeon}.} \bibinfo{year}{2020}\natexlab{}.
\newblock \showarticletitle{Untact: Customer’s acceptance intention toward
  robot barista in coffee shop}.
\newblock \bibinfo{journal}{\emph{Sustainability}} \bibinfo{volume}{12},
  \bibinfo{number}{20} (\bibinfo{year}{2020}), \bibinfo{pages}{8598}.
\newblock


\bibitem[\protect\citeauthoryear{ter Hoeven, Kwakkel, Hess, Pike, Wang, rht,
  and Kazil}{ter Hoeven et~al\mbox{.}}{2025}]%
        {terHoeven2025}
\bibfield{author}{\bibinfo{person}{Ewout ter Hoeven}, \bibinfo{person}{Jan
  Kwakkel}, \bibinfo{person}{Vincent Hess}, \bibinfo{person}{Thomas Pike},
  \bibinfo{person}{Boyu Wang}, \bibinfo{person}{rht}, {and}
  \bibinfo{person}{Jackie Kazil}.} \bibinfo{year}{2025}\natexlab{}.
\newblock \showarticletitle{Mesa 3: Agent-based modeling with Python in 2025}.
\newblock \bibinfo{journal}{\emph{Journal of Open Source Software}}
  \bibinfo{volume}{10}, \bibinfo{number}{107} (\bibinfo{date}{March}
  \bibinfo{year}{2025}), \bibinfo{pages}{7668}.
\newblock
\showISSN{2475-9066}
\urldef\tempurl%
\url{https://doi.org/10.21105/joss.07668}
\showDOI{\tempurl}


\bibitem[\protect\citeauthoryear{Torvald}{Torvald}{2002}]%
        {yarchiveEverythingisafilePrinciple}
\bibfield{author}{\bibinfo{person}{Linus Torvald}.}
  \bibinfo{year}{2002}\natexlab{}.
\newblock \bibinfo{title}{{T}he everything-is-a-file principle ({L}inus
  {T}orvalds) --- yarchive.net}.
\newblock
  \bibinfo{howpublished}{\url{https://yarchive.net/comp/linux/everything_is_file.html}}.
\newblock
\newblock
\shownote{[Accessed 22-09-2025].}


\bibitem[\protect\citeauthoryear{Towers, Kwiatkowski, Terry, Balis, De~Cola,
  Deleu, Goul{\~a}o, Kallinteris, Krimmel, KG, et~al\mbox{.}}{Towers
  et~al\mbox{.}}{2024}]%
        {towers2024gymnasium}
\bibfield{author}{\bibinfo{person}{Mark Towers}, \bibinfo{person}{Ariel
  Kwiatkowski}, \bibinfo{person}{Jordan Terry}, \bibinfo{person}{John~U Balis},
  \bibinfo{person}{Gianluca De~Cola}, \bibinfo{person}{Tristan Deleu},
  \bibinfo{person}{Manuel Goul{\~a}o}, \bibinfo{person}{Andreas Kallinteris},
  \bibinfo{person}{Markus Krimmel}, \bibinfo{person}{Arjun KG},
  {et~al\mbox{.}}} \bibinfo{year}{2024}\natexlab{}.
\newblock \showarticletitle{Gymnasium: A standard interface for reinforcement
  learning environments}.
\newblock \bibinfo{journal}{\emph{arXiv preprint arXiv:2407.17032}}
  (\bibinfo{year}{2024}).
\newblock


\bibitem[\protect\citeauthoryear{Truong, Rudolph, Yokoyama, Chernova, Batra,
  and Rai}{Truong et~al\mbox{.}}{2023}]%
        {truong2023rethinking}
\bibfield{author}{\bibinfo{person}{Joanne Truong}, \bibinfo{person}{Max
  Rudolph}, \bibinfo{person}{Naoki~Harrison Yokoyama}, \bibinfo{person}{Sonia
  Chernova}, \bibinfo{person}{Dhruv Batra}, {and} \bibinfo{person}{Akshara
  Rai}.} \bibinfo{year}{2023}\natexlab{}.
\newblock \showarticletitle{Rethinking sim2real: Lower fidelity simulation
  leads to higher sim2real transfer in navigation}. In
  \bibinfo{booktitle}{\emph{conference on Robot Learning}}. PMLR,
  \bibinfo{pages}{859--870}.
\newblock


\bibitem[\protect\citeauthoryear{Wilensky}{Wilensky}{1999}]%
        {Wilensky:1999}
\bibfield{author}{\bibinfo{person}{U. Wilensky}.}
  \bibinfo{year}{1999}\natexlab{}.
\newblock \bibinfo{booktitle}{\emph{NetLogo}}.
\newblock \bibinfo{type}{http://ccl.northwestern.edu/netlogo/}.
  \bibinfo{institution}{Center for Connected Learning and Computer-Based
  Modeling}, \bibinfo{address}{Northwestern University, Evanston, IL}.
\newblock
\urldef\tempurl%
\url{http://ccl.northwestern.edu/netlogo/}
\showURL{%
\tempurl}


\end{thebibliography}

%%%%%%%%%%%%%%%%%%%%%%%%%%%%%%%%%%%%%%%%%%%%%%%%%%%%%%%%%%%%%%%%%%%%%%%%

\end{document}